# Robotic Hand Rehabilitation System

Dineshkumar V
Department of Electronics and
Communication Engineering
Amrita School of Engineering, Coimbatore
Amrita Vishwa Vidyapeetham, India

N R Sakthivel
Department of Mechanical Engineering
Amrita School of Engineering, Coimbatore
Amrita Vishwa Vidyapeetham, India

Binoy B Nair
Department of Electronics and
Communication Engineering
Amrita School of Engineering, Coimbatore
Amrita Vishwa Vidyapeetham, India

**Abstract**
Rehabilitation exercises are essential to ensure speedy recovery of stroke patients. An automated system to assist the patient in performing a rehabilitation exercise repeatedly is designed. The design process is presented in this report.

## 1. Introduction

Recovery of stroke victims depends to a large extent on the regularity with which the patient performs the rehabilitation exercises. Robotic systems are useful for the purpose. The design of such a system to help the stroke patients exercise is presented here. The system uses stepper motors to develop a robotic rehabilitation device. Details of the system are presented below. The block diagram of the robotic rehabilitation device is presented in Figure 1.

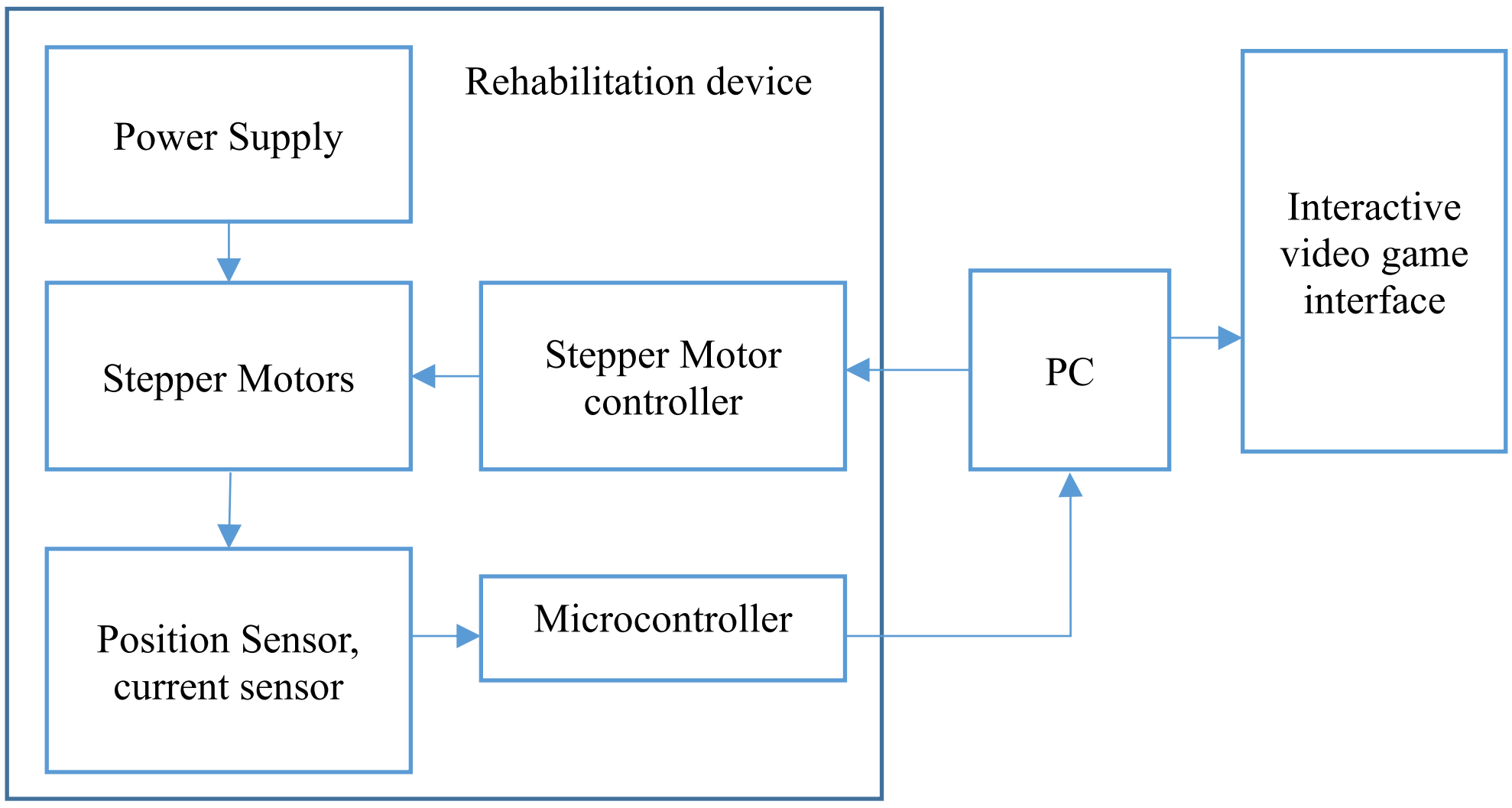

Figure 1 Robotic Hand exercise system block diagram

## 2. Simulation Studies

The stepper motor is sourced from National Instruments (Part Number N33HRLG-LEK-M2-00)[1]. The motor specification are as follows:

Table 1 Motor Parameters

| S. No. | Parameter | Value |
|---|---|---|
| 1 | Step angle | 1.8 degrees/step |
| 2 | Steps per revolution | 200 |
| 3 | Angular accuracy | ±3% |
| 4 | Phases | 2 |
| 5 | Amps/Phase (A) | 1.24 |
| 6 | Holding Torque oz-in. (N . m) | 1710 (12.08) |
| 7 | Rotor Inertia oz-in.-s2 (kg-m2 x10-3) | 0.0567 (0.4) |
| 8 | Phase Inductance (mH) | 144 |
| 9 | Phase Resistance Ω ±10% | 13 |
| 10 | Detent Torque oz-in. (N . m) | 54 (0.381) |
| 11 | Thermal Resistance °C/watt | 1.61 |
| 12 | Maximum RPM | 1800 |

The motor torque constant is calculated using the equation:

$$K_T = \frac{K_H}{I_s} \qquad (1)$$

Where $K_T$ = Torque constant , $K_H$= Holding Torque and $I_s$ = stator current
Therefore, for this motor, the torque constant $K_T$ = 12.08/1.24 =9.74

The simulation of Stepper motor and control is carried out using MATLAB Simscape[2]. The control strategy was developed and realized using MATLAB Simulink. The system is designed to accept two inputs: the duration of the exercise and the number of steps. The control system works by rotating the shaft of the motor by the specified number of steps at the specified rate till it reaches its final state. The position is held for a fixed duration and then the reverse motion starts in the opposite direction till the same number of steps are taken and the motor shaft comes back to rest in its initial position. The Simulink model for the control algorithm is presented in Figure 2. The top-level diagram is given in Figure 2 (a).
The duration of the cycle is the input to the algorithm which has been realized in the 'Stepper Controller' block. The output of this block is a PWM signal that is used to drive the stepper motor driver which produces the stepper motor control signals. It creates the pulse trains, A and B, required to control the motor. This block initiates a step each time the voltage at the ENA port rises above the Enable threshold voltage parameter value.

In the actual implementation we work with TTL logic voltage levels, i.e. 0V and 5V , hence the threshold is set at 2.5V. If the voltage at the REV port is less than or equal to the Reverse threshold voltage parameter value (again, set at 2.5V), pulse A leads pulse B by 90 degrees. If the voltage at the REV port is greater than the Reverse threshold voltage value, pulse B leads pulse A by 90 degrees and the motor direction is reversed.

The Stepper Motor block represents a stepper motor. It uses the input pulse trains, A and B, to control the mechanical output according to the following equations:

$$e_A = -K_m\omega\, sin(N_r\theta) \qquad (2)$$

$$e_B = K_m\omega\, cos(N_r\theta) \qquad (3)$$

$$\frac{di_A}{dt} = \frac{v_A - Ri_A - e_A}{L} \qquad (4)$$

$$\frac{di_B}{dt} = \frac{v_B - Ri_B - e_B}{L} \qquad (5)$$

$$J\frac{d\omega}{dt} + B\omega = T_e \qquad (6)$$

$$T_e = -K_m(i_A - \frac{e_A}{R_m})sin(N_r\theta) + K_m(i_B - \frac{e_B}{R_m})cos(N_r\theta) - T_d sin(4N_r\theta) \qquad (7)$$

$$\frac{d\theta}{dt} = \omega \qquad (8)$$

where:

$e_A$ and $e_B$ are the back electromotive forces (emfs) induced in the A and B phase windings, respectively.
$i_A$ and $i_B$ are the A and B phase winding currents.
$v_A$ and $v_B$ are the A and B phase winding voltages.
$K_m$ is the motor torque constant.
$N_r$ is the number of teeth on each of the two rotor poles. The step size is $(\pi/2)/N_r$.
$R$ is the winding resistance.
$L$ is the winding inductance.
$R_m$ is the magnetizing resistance.
$B$ is the rotational damping.
$J$ is the inertia.
$\omega$ is the rotor speed.
$\Theta$ is the rotor angle.
$T_d$ is the detent torque amplitude.
$T_e$ is the electrical torque.

If the initial rotor is zero or some multiple of $(\pi/2)/N_r$, the rotor is aligned with the phase winding of pulse A. This happens when there is a positive current flowing from the A+ to the A- ports and there is no current flowing from the B+ to the B- ports. The Stepper Motor block produces a positive torque acting from the mechanical C to R ports when the phase of pulse A leads the phase of pulse B. In this study, the $R_m$ is considered to be a very large number leading to ignoring the fractions $\frac{e_A}{R_m}$ and $\frac{e_B}{R_m}$ in equation (7).

The internal structure of the stepper controller is presented in detail in Figure 2(b), (c), (d) and (e). Figure 2(b) and (c) represent the voltage level conversion to 0-5V voltage levels and conversion of the signal types for interoperability with Simscape stepper motor model. The PWM pulse generation process is detailed in Figure 2 (d) and (e).

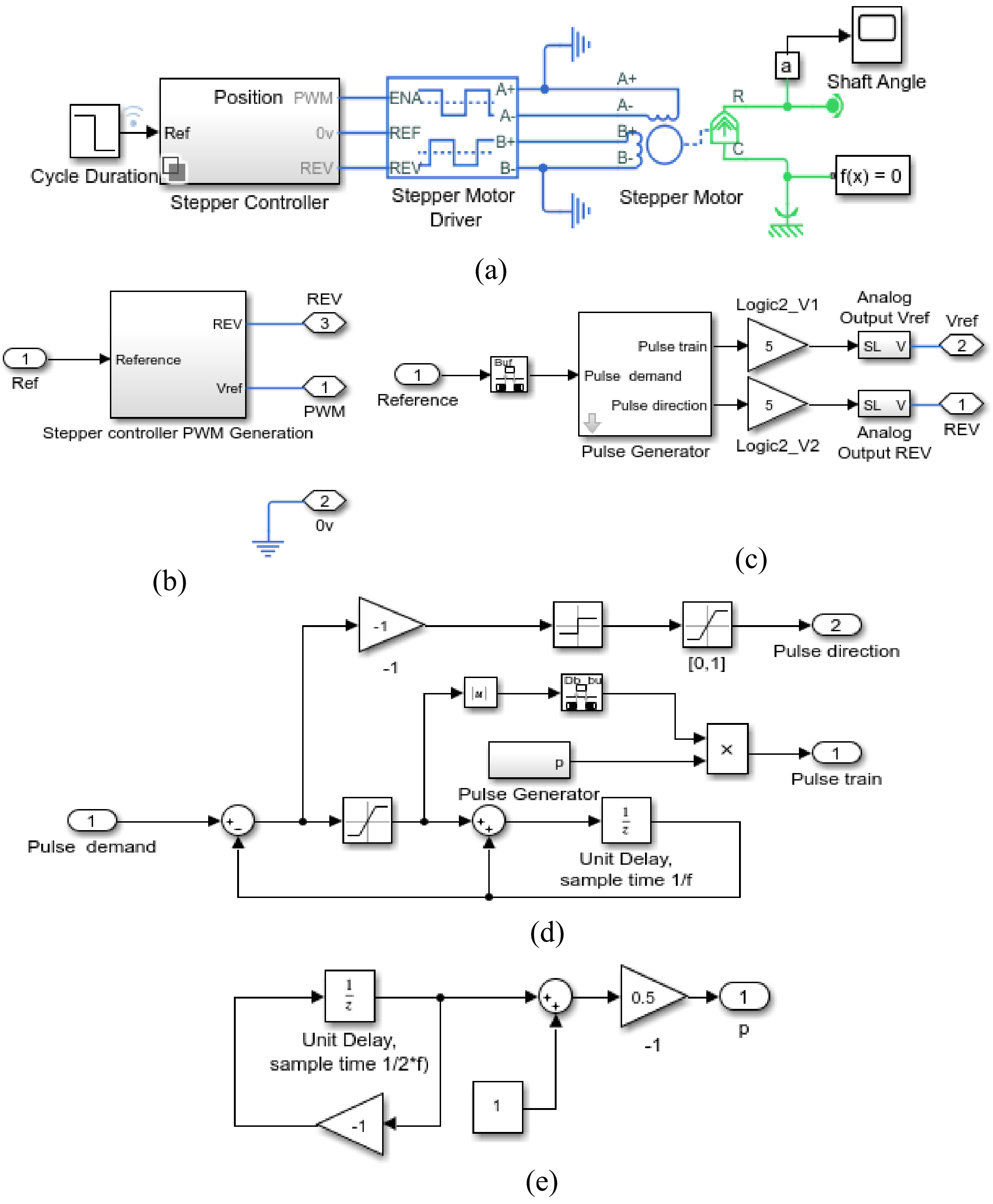

Figure 2 Stepper motor with control algorithm

The simulation output of the system is presented in Figure 3. Simulation studies for various movement angles were carried out. The Figure 3 below shows movement of 20 steps (20*1.8 =36 degrees).

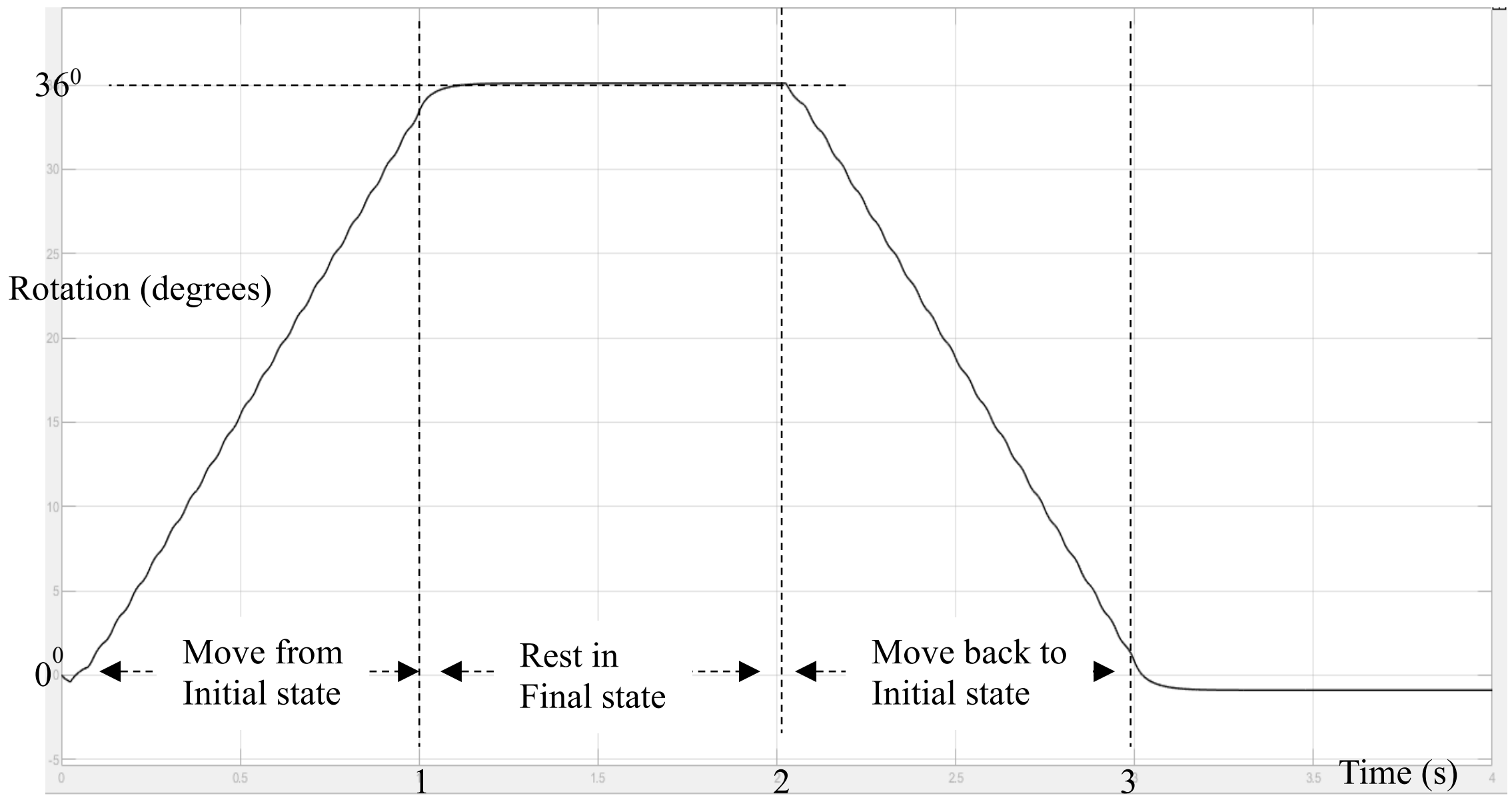

Figure 3 Stepper motor shaft angle vs time

The stepper motor control signals and the current waveforms under no-load condition are presented in Figure 4.

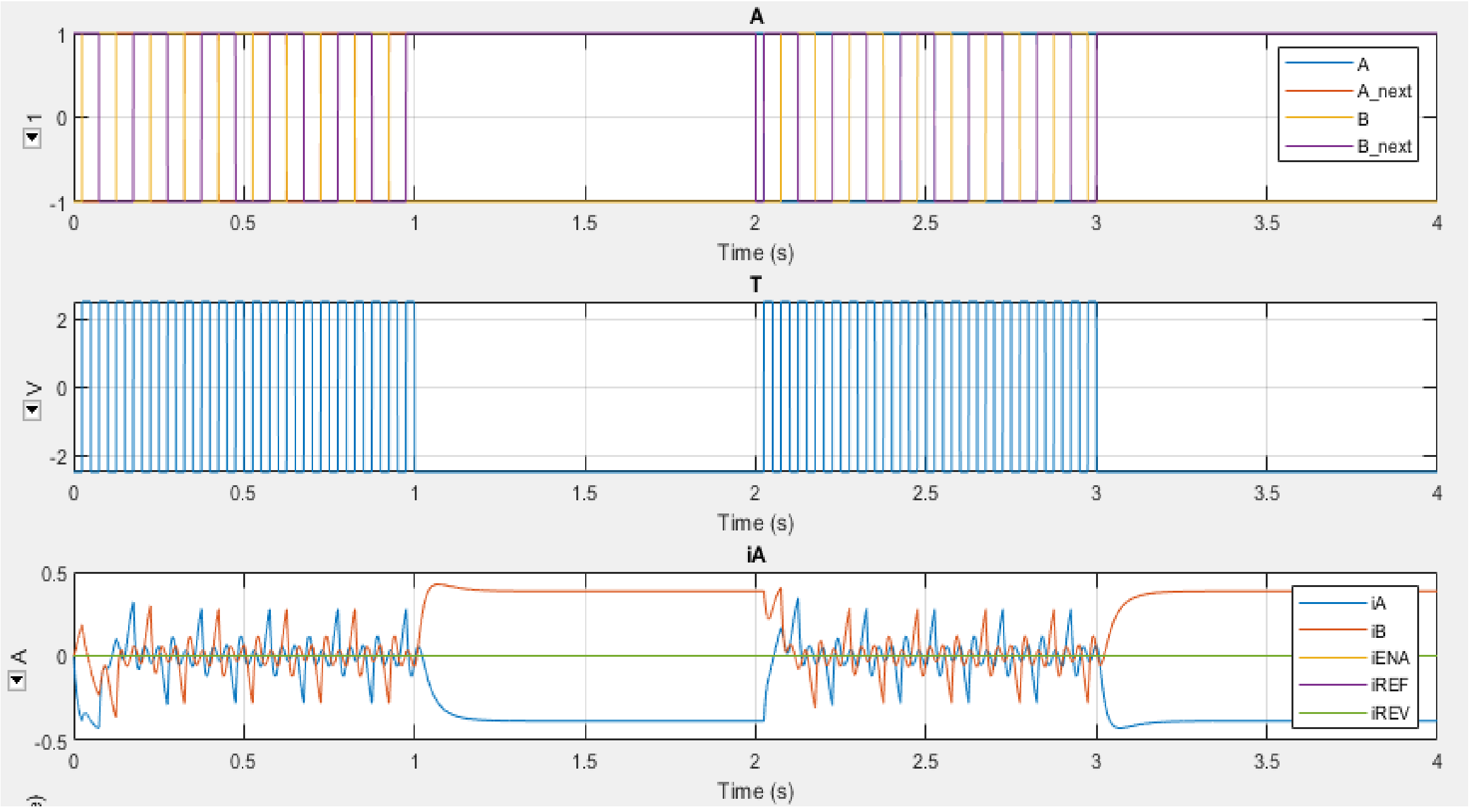

Figure 4 Stepper motor control signals and current waveforms

The mechanical system simulation model using Simscape is presented in Figure 5. The complete mechanical system model is presented in Figure 5(a). The subsystem model within the Rigid Transform 3- Rigid Transform 4 block is presented in Figure 5(b). A revolute joint acts as the point where the input from the stepper motor is received and the rigid body connected to the revolute joint moves accordingly. The material considered as the rigid body is mild-steel with density 7850 $kg/m^3$. The length of the rigid body is considered to be of length 1m and width 2cm and height of 1cm.

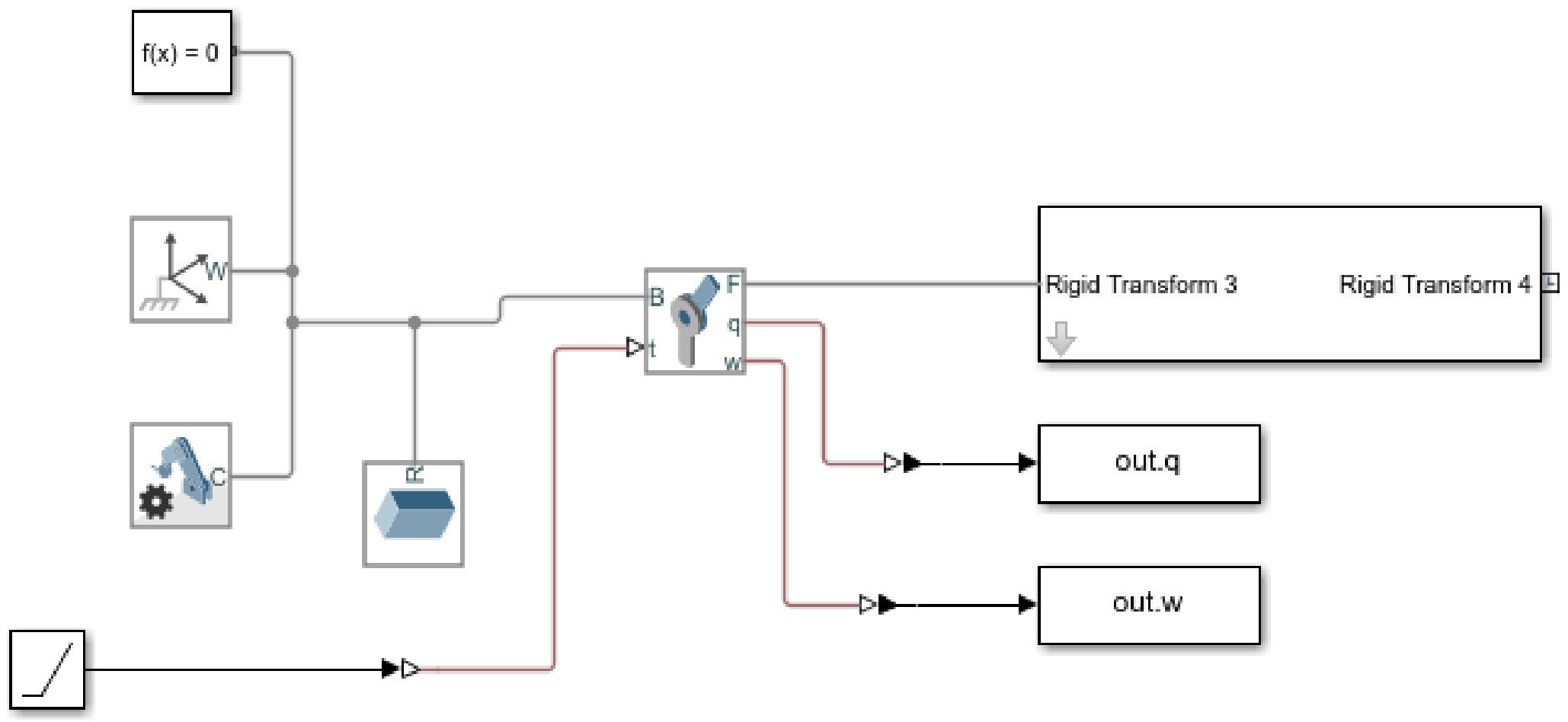

(a)

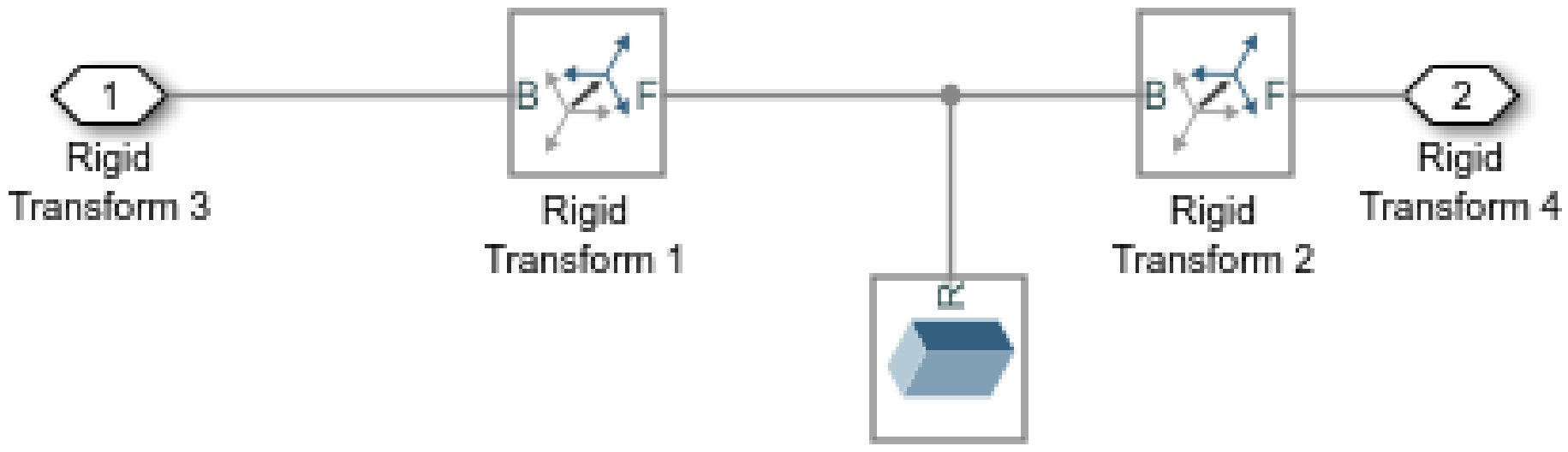

(b)

Figure 5 Mechanical system simulation model

Simulation results for the Simscape mechanical model is presented in Figure 6. The initial position is presented in Figure 6(a) and the final position in Figure 6(b).

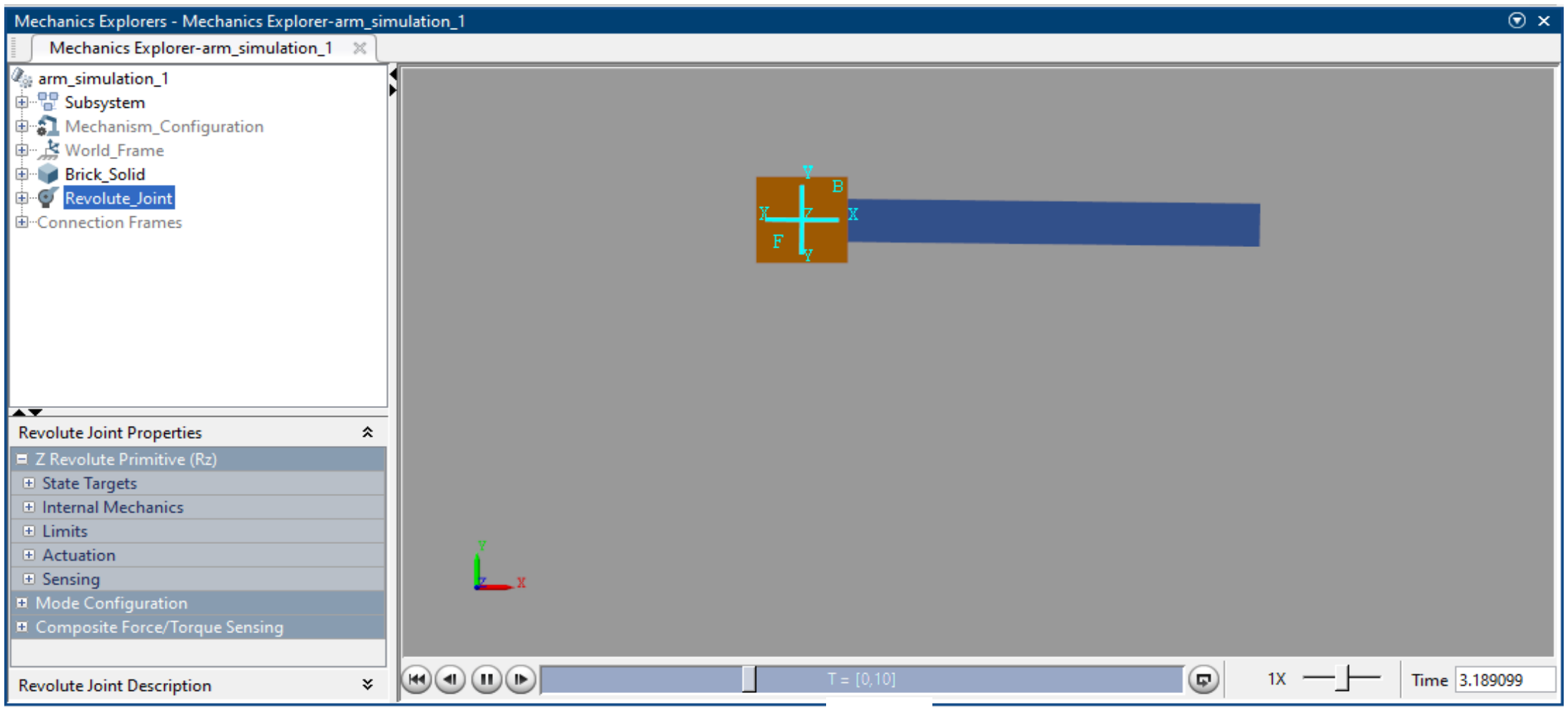

(a)

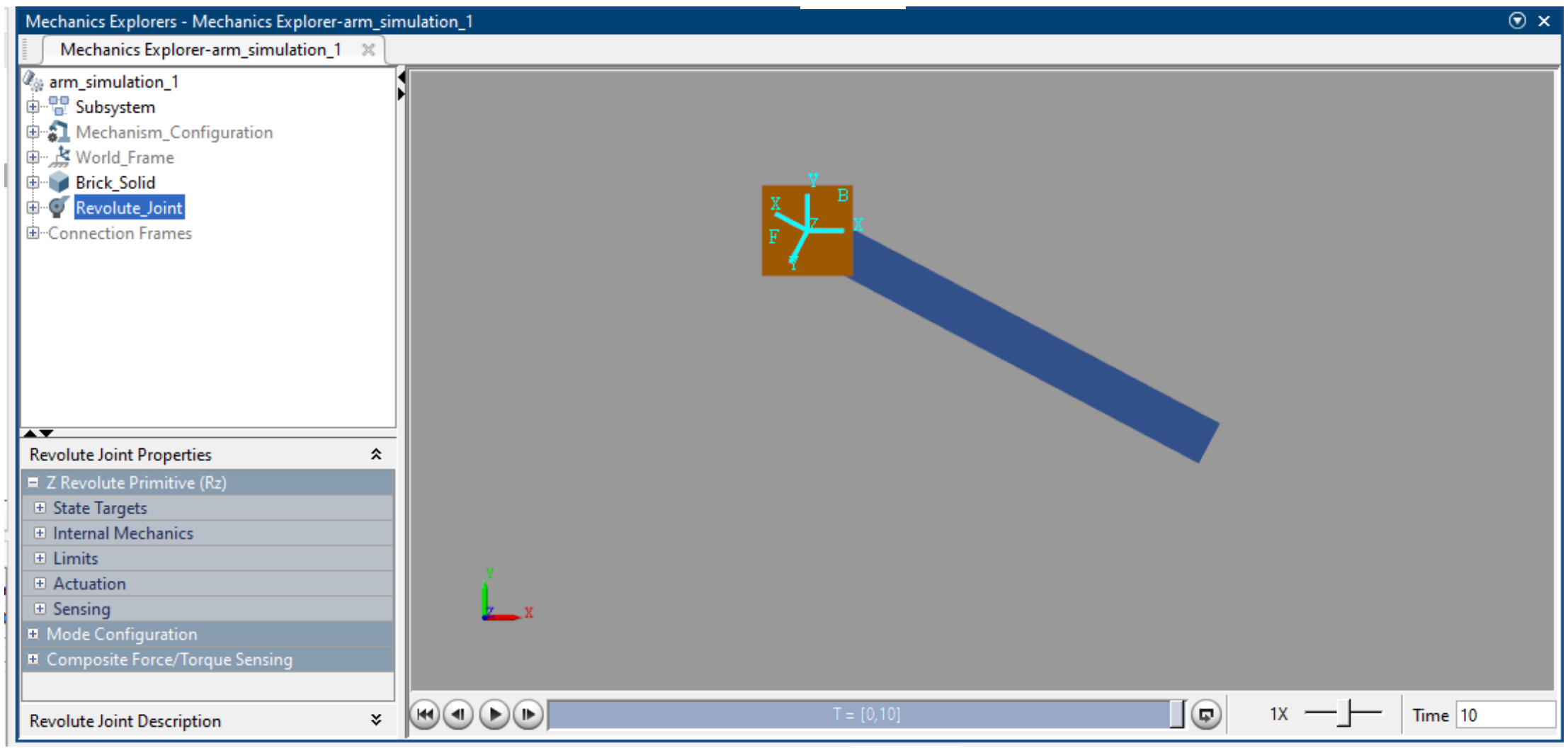

(b)

Figure 6 Mechanical system simulation results (a) initial position and (b) final position

The movement of the mechanical arm model is presented in Figure 7. As can be seen, the model output indicates smooth movement from the initial to the final state. The y-axis is in radians.

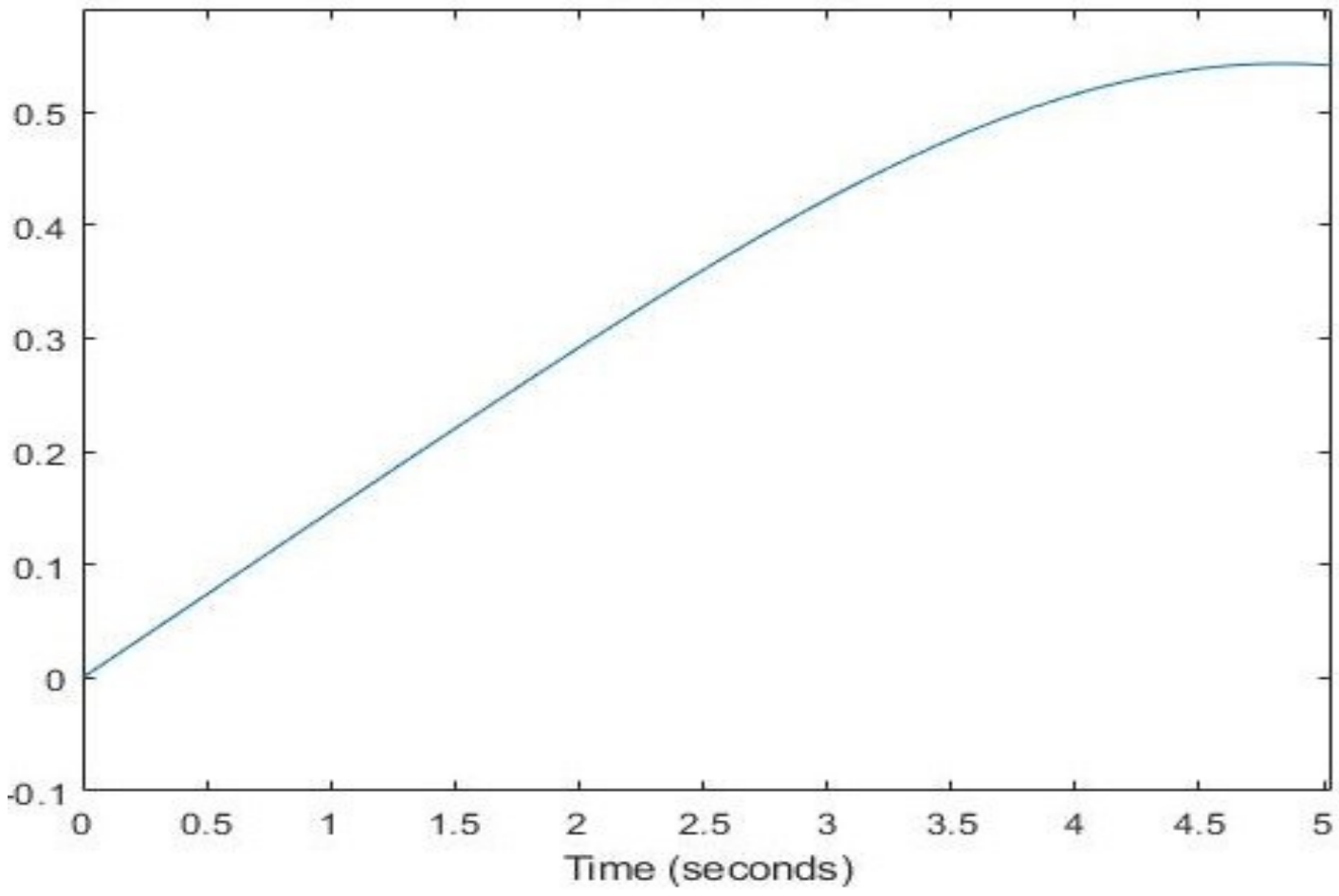

Figure 7 Mechanical arm model position over time

## 3. Hardware Setup

Measurement and Automation Explorer, NI Motion Assistant[3] and NI LabVIEW Software are required to control the Stepper motors.

1. In NI-MAX, settings on how to configure the axis and stepper settings such as stepper polarity, steps per revolution of the motor shaft as well as direction of the motor run are configured.
2. In Motion Input/ Output settings, options on how to configure the home switches, Limit switches (forward switches and Reverse switches) software switches and inhibits on the motion system are present. With the help of this, the active state in our I/O to enable/ disable and active low/ high polarity can be set. To configure the movement of the motor, units such as Velocity, Acceleration and Deceleration have to be provided.
3. Using find reference settings, we would configure the reference moves, in order to find the home switch (the initial search direction, final approach direction, edge of the switch to stop on and approach velocity and set the offset move, after the home is found, and to determine the reset position after the offset move.
4. After initialization, with the help of interactive environment that allows testing and debugging the system, the motor function can be tested.
5. Following parameters need to be provided: axis to run, stepper loop mode, operation mode, target position of axis. Now the axis position, axis velocity and axis velocity & position with respect to axis can be plotted.
6. After the Initialization, motion applications can be developed with help of NI Motion Assistant.
7. Motion Assistant is used to graphically construct, preview and test motion application. After that, the LabVIEW VI can be generated.

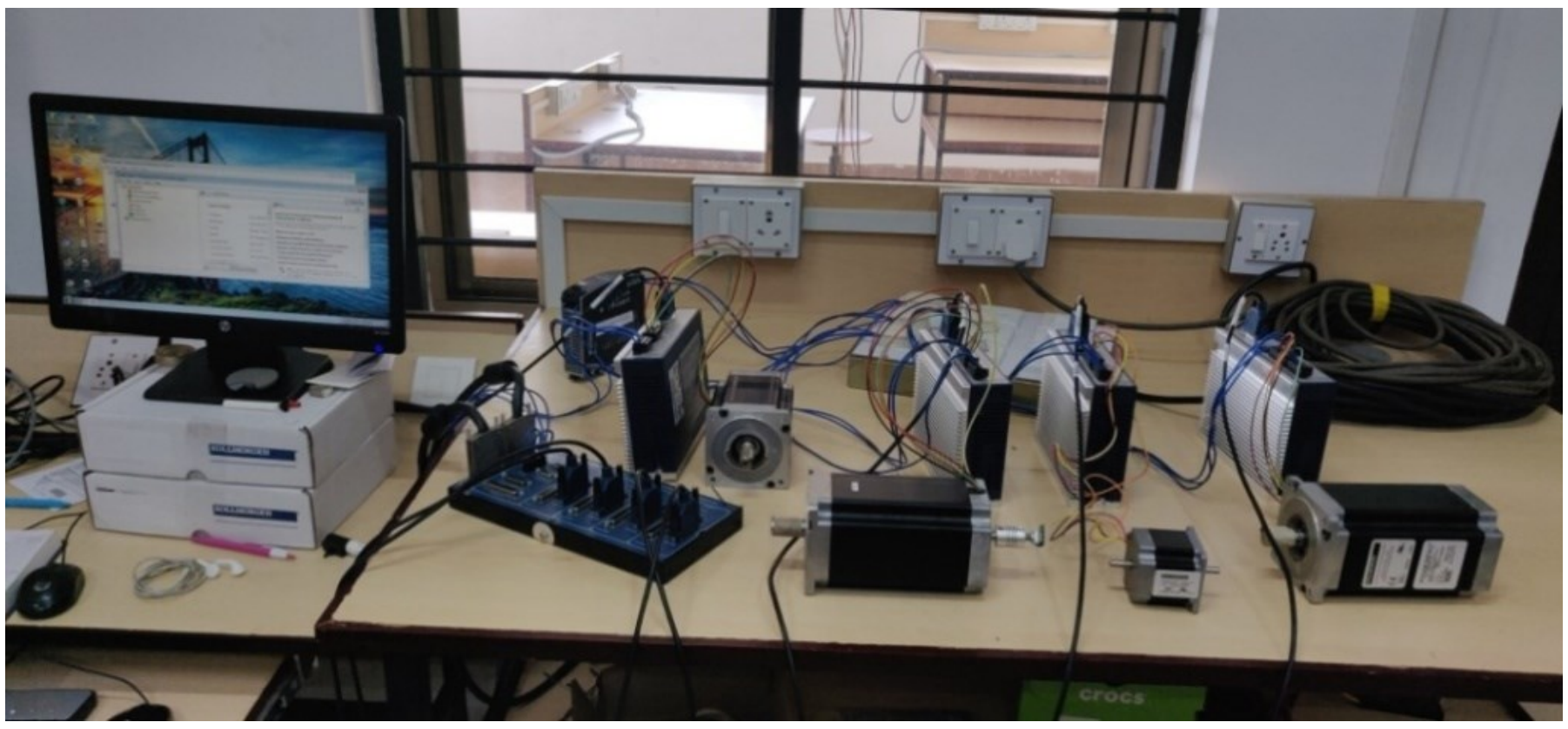

Figure 8 Stepper motor bundle hardware setup

### 3.1 Measurement and Automation Explorer:

3.1.1 Axis configuration:

1. Axis type: sets the axes to servomotor or stepper motor.
2. Axis enabled : enables/ disables the axes

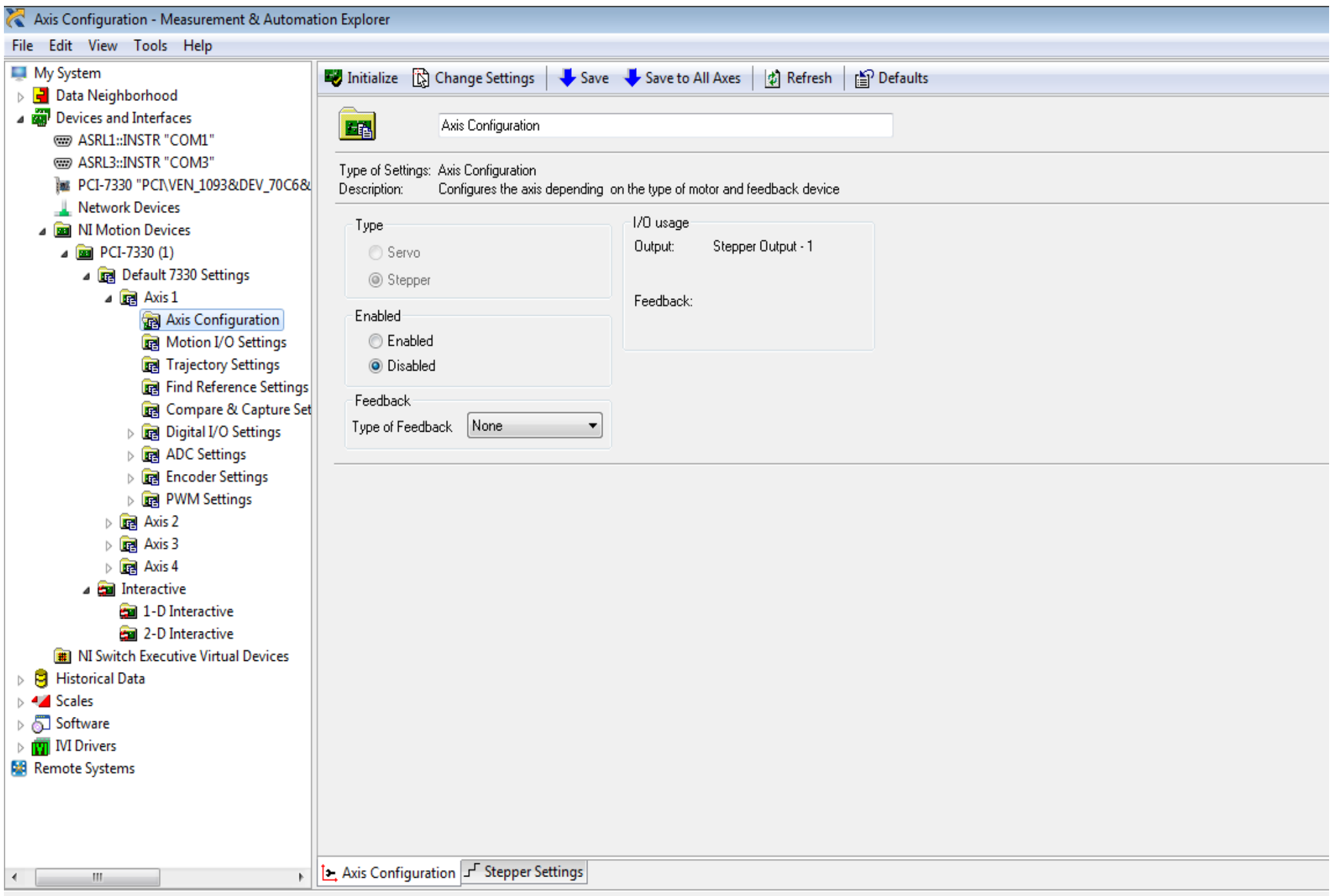

Figure 9 Axis configuration (a)

Configuration settings 1 and 2 are presented in Figure 9

3. Stepper resolution: set the resolution for the feedback and stepper motor (micro stepping)
4. Stepper specific settings – sets the nature of the stepper motor.
5. Stepper loop mode: open or closed loop.
6. Stepper polarity : Active high or active low
7. Stepper output mode: step and direction or clockwise or anti clock direction.

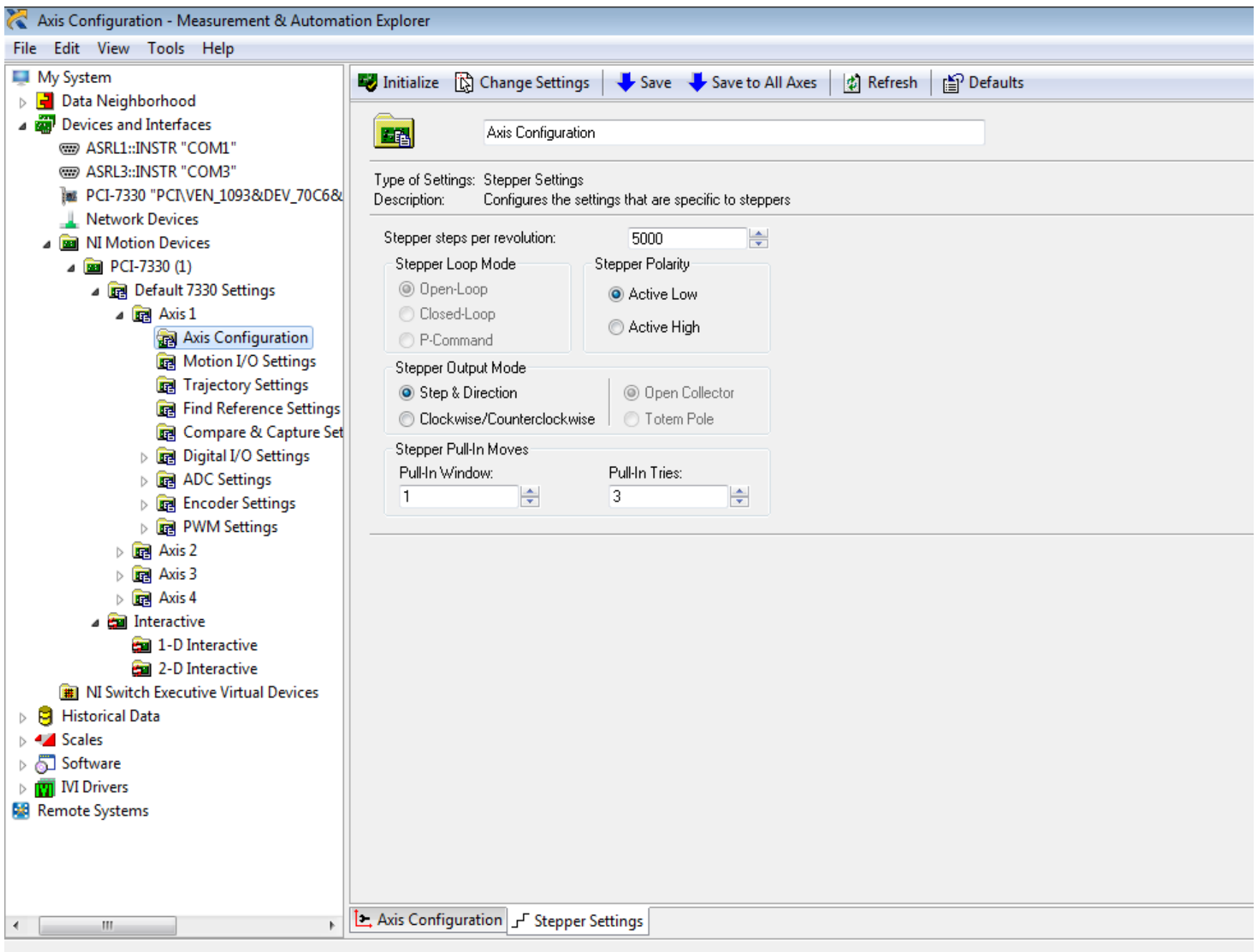

Figure 10 Axis configuration (b)

Configuration settings 3- 7 are presented in Figure 10.

3.1.2 Motion settings:

The motion I/O tab can be used to configure the limit and home switches. The active state in our I/O can be set to enable/ disable and active low/ high polarity.

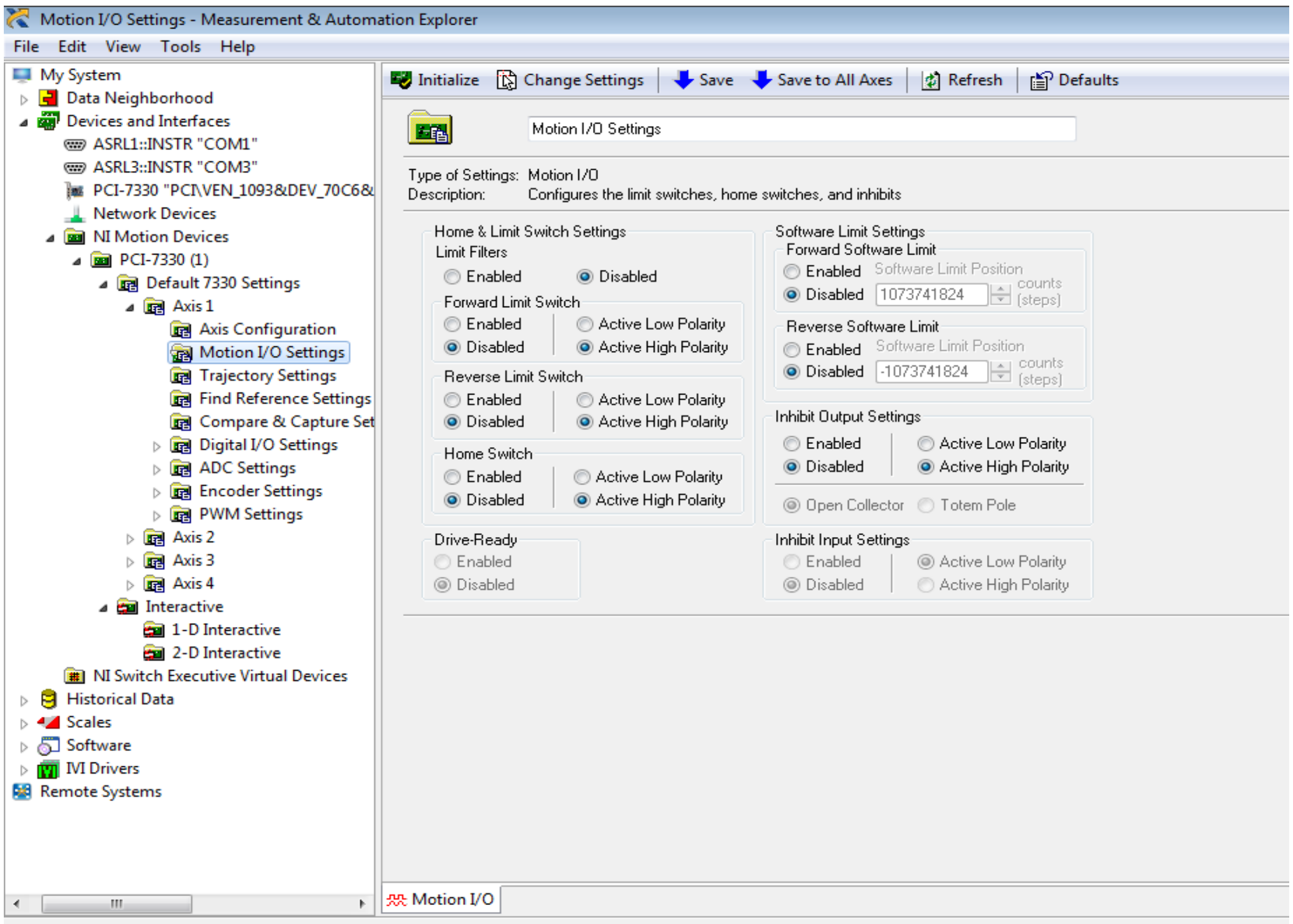

Figure 11 Motion settings

3.1.3 Interactive settings

MAX for motion control offers 1D and 2D interactive panels that allow testing the motion system after the configuration is done. These panels are used to view the graphs of the position and the velocity of the moves

1D INTERACTIVE -Used to test the setup performance of any one axis at a time.

2D INTERACTIVE- Used to test the setup performance of any one axis at a time. The position and velocity graphs on the position and Misc. plots tabs can be used to view the motion.

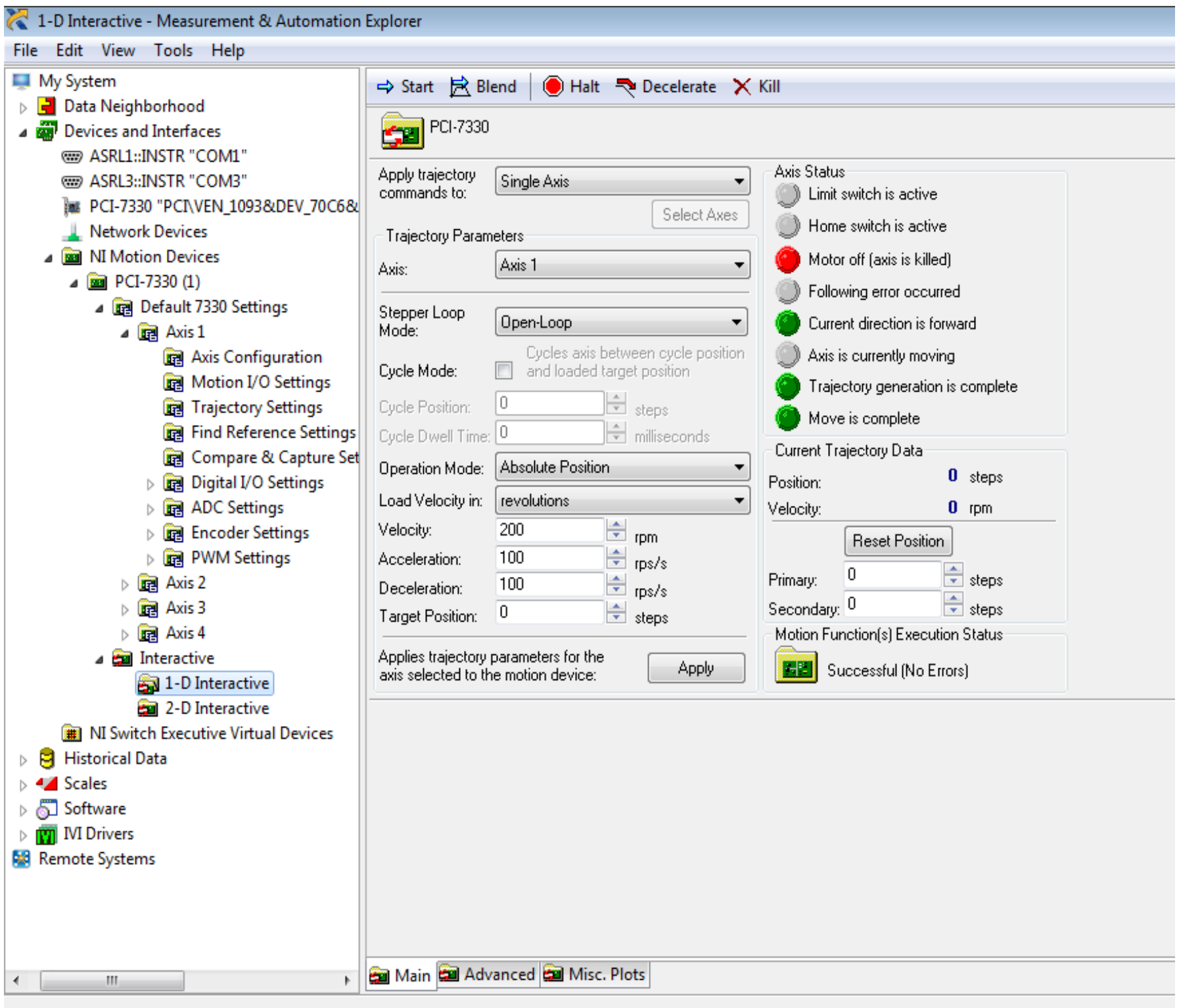

Figure 12 1D interactive settings

## 3.2 NI Motion Assistant

NI Motion Assistant is a stand-alone prototyping tool for quickly developing motion applications. NI Motion Assistant allows graphical construction and preview motion applications without writing any code. When a prototype in NI Motion Assistant is completed, LabVIEW or C code can be generated for further application development. Using this, various types of movements can be done:

1. Straight move – point A to point B movement. Based on the velocity or position.
2. Arc move - point A to point B in arc or helix. Based on the velocity or position.
3. Contoured move- user defined move, to generate the trajectory and the points loaded into the motion controller are splined to create a smooth profile.
4. Reference move – initialize the axes to known physical reference as a home switches.

### 3.2.1 Move profiles:

The basic function of motion controller is to make moves. The trajectory generates takes in the type of move and move constraints and generates point in a real time.

Trapezoidal: Set the acceleration to specified value and then move at the maximum velocity set. The velocity of the axes in a coordinate space never exceed the maximum velocity loaded. The axes decelerates to a stop at their final position.

S-curve: acceleration and deceleration are smooth, resulting in less abrupt transitions. This limits the jerk in the motion control system but increases cycle time. The value by which the profile is smoothed is called the maximum jerk or s-curve value. The Lab view algorithm based on the type of movements:

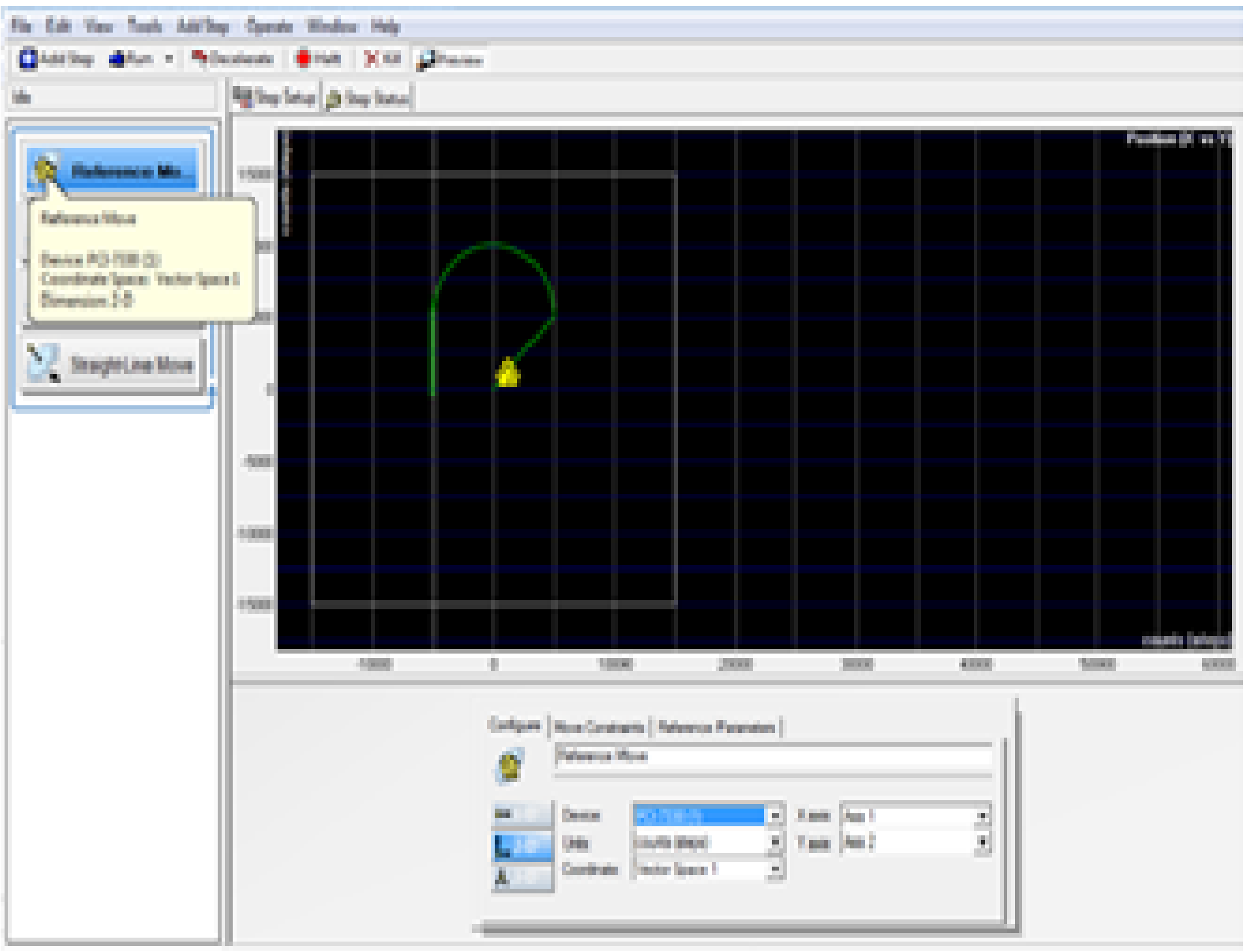

Figure 13 NI Motion Assistant window

### 3.3 LabVIEW VI

1. Operation mode :

   Axes : To select the axes. We have four axes. Choose which axes of the motor to run.

   Motion : Absolute motion or Relative motion
2. Load the target position and velocity of the motor.
3. Load acceleration and deceleration of the motor.
4. Read the value of position and velocity while running.
5. Whether the target position is reached or not. It will show our task is completed or not.
6. Stop the motor, when some disturbance occurs.

The VI for the system is given in Figure 14.

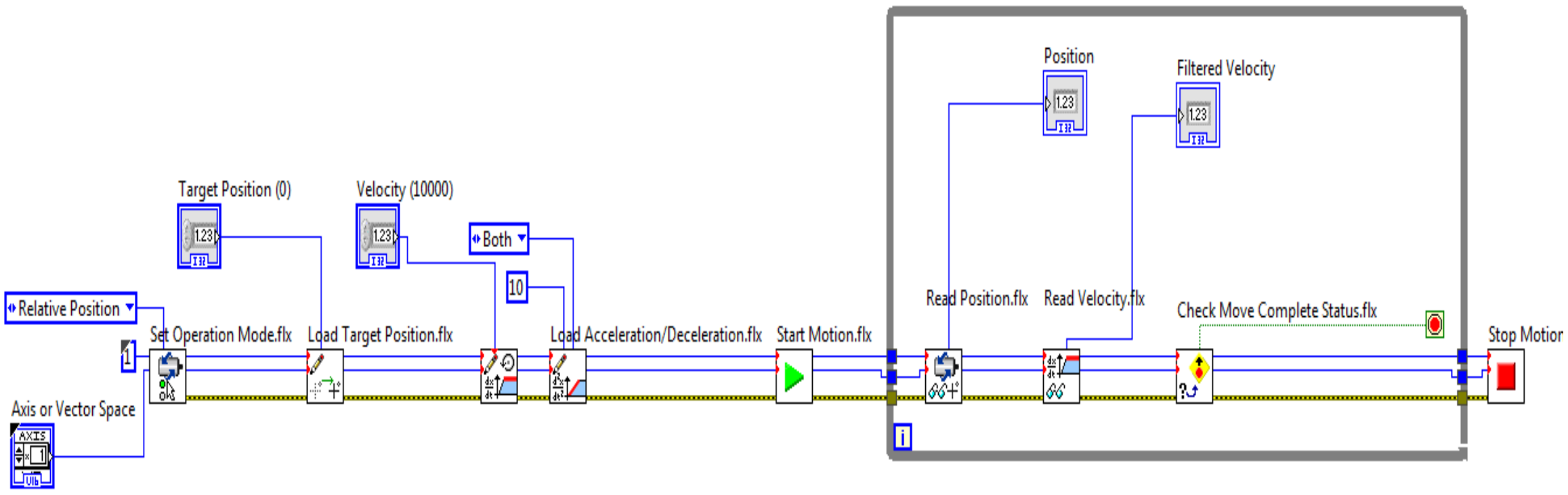

Figure 14 LabVIEW VI to run the motor

The overall CAD diagram of the seating arrangement and the mounting frame are given in figure below.

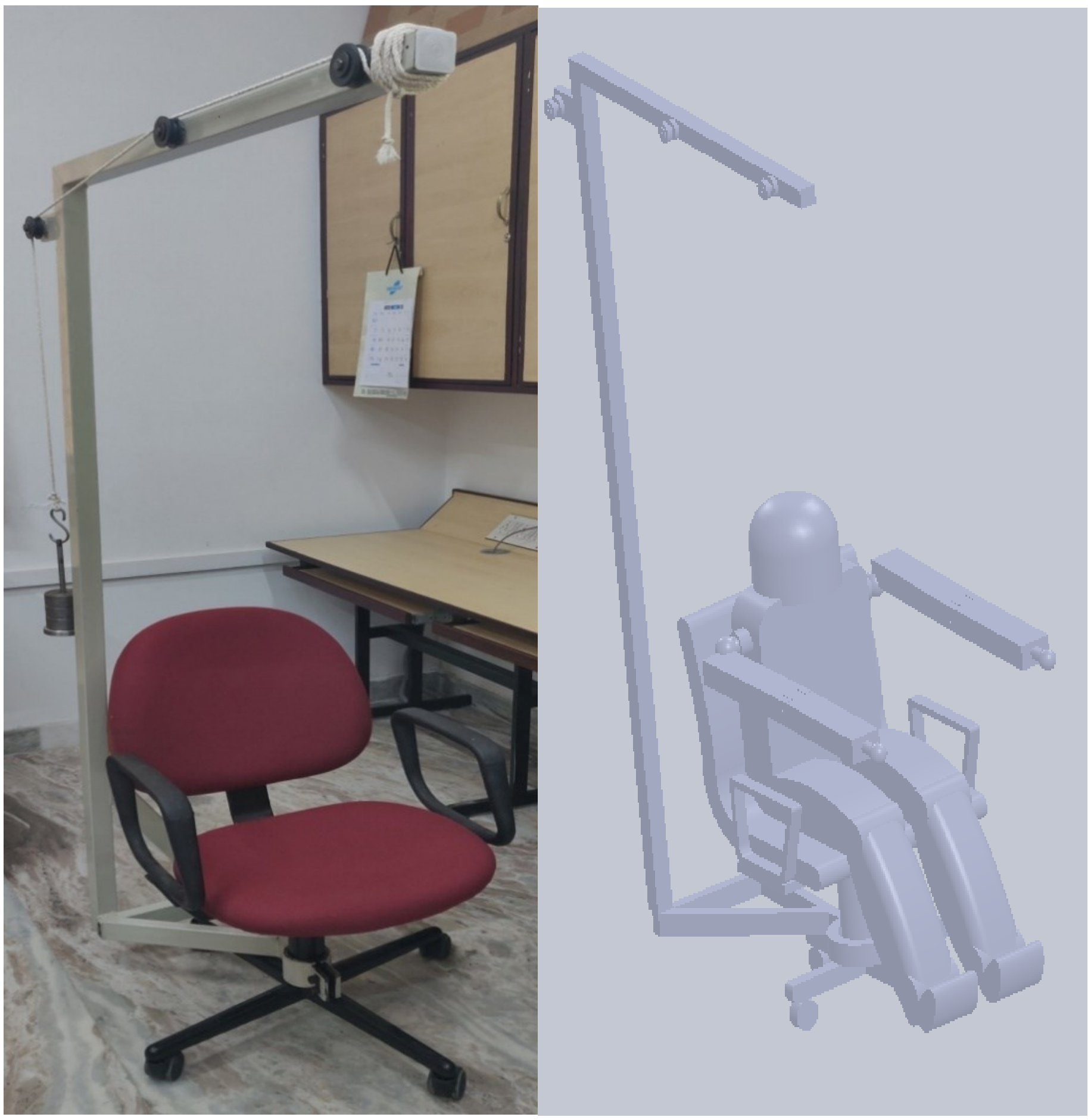

Figure 15 Prototype and the CAD design

Funding: The system presented in this report has been developed using funds received from Department of Science and Technology, India, under Technology intervention for disabled and elderly (TIDE) scheme. The authors gratefully acknowledge the financial support provided by DST.